\documentclass[10pt,twocolumn,letterpaper]{article}

\usepackage{iccv}
\usepackage{times}
\usepackage{epsfig}
\usepackage{graphicx}
\usepackage{amsmath}
\usepackage{amssymb}
\usepackage{subfig}
\usepackage{float}

\usepackage{booktabs}
\usepackage[pagebackref=true,breaklinks=true,letterpaper=true,colorlinks,bookmarks=false]{hyperref}

\usepackage{color}

\makeatletter
\renewcommand{\paragraph}{%
  \@startsection{paragraph}{4}%
  {\z@}{0.5ex \@plus 1ex \@minus .2ex}{-1em}%
  {\normalfont\normalsize\bfseries}%
}
\makeatother

\iccvfinalcopy % *** Uncomment this line for the final submission

\def\iccvPaperID{9359} % *** Enter the ICCV Paper ID here

\ificcvfinal\fi

\begin{document}

%%%%%%%%% TITLE
\title{Image-to-Image Translation with Low Resolution Conditioning}

\author{Mohamed Abid$^{*}$, Ihsen Hedhli$^{*}$, Jean-François Lalonde$^{*}$, Christian Gagne$^{* \dagger}$ \\
% {\tt\small Mohamed-abderrahmen-abid.1@ulaval.ca }\\, {\tt\small ihsen.hedhli@iid.ulaval.ca}\\
% {\tt\small$\left\{$jflalonde, christian.gagne$\}\right$ @gel.ulaval.ca }\\ 
$^{*}$Université Laval, $^{\dagger}$Canada CIFAR AI Chair, Mila
%\\
}

\maketitle
\ificcvfinal\thispagestyle{empty}\fi

% List of sections
%!TEX root = egbib.tex
\begin{abstract}
% Image-to-image translation method is an important domain of computer vision and has been applied to various applications. 
Most image-to-image translation methods focus on learning mappings across domains with the assumption that images share content (e.g., pose) but have their own domain-specific information known as style. When conditioned on a target image, such methods aim to extract the style of the target and combine it with the content of the source image. 
In this work, we consider the scenario where the target image has a very low resolution. More specifically, our approach aims at transferring fine details from a high resolution (HR) source image to fit a coarse, low resolution (LR) image representation of the target. We therefore generate HR images that share features from both HR and LR inputs. This differs from previous methods that focus on translating a given image style into a target content, our translation approach being able to simultaneously imitate the style and merge the structural information of the LR target. Our approach relies on training the generative model to produce HR target images that both 1) share distinctive information of the associated source image; 2) correctly match the LR target image when downscaled. We validate our method on the CelebA-HQ and AFHQ datasets by demonstrating improvements in terms of visual quality, diversity and coverage. Qualitative and quantitative results show that when dealing with intra-domain image translation, our method generates more realistic samples compared to state-of-the-art methods such as Stargan~v2~\cite{starganv}.
\end{abstract}
%!TEX root = egbib.tex
\section{Introduction}

\begin{figure}[ht]
\begin{center}
%\fbox{\rule{0pt}{2in} \rule{.9\linewidth}{0pt}}
\includegraphics[width=\linewidth]{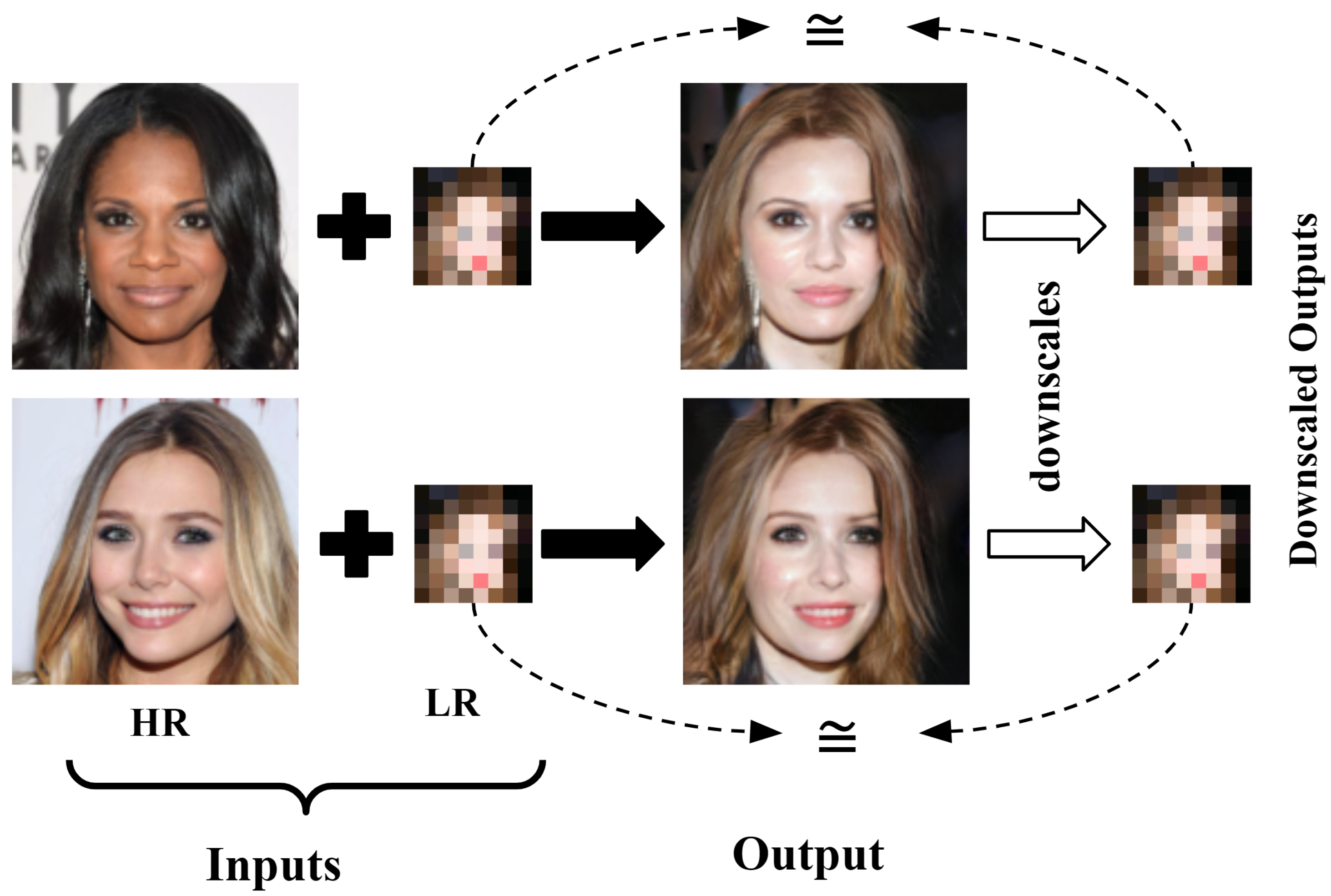}
\end{center}
   \caption{Illustration of the proposed approach on CelebA-HQ dataset: our method learned to map the inputs HR image (source) and LR (target) image to generate a HR output that preserves the identity of the source image and stays faithful to the structure of the LR target.}
\label{fig:first}
\end{figure}

Image-to-image translation (i2i) methods seek to learn a mapping between two related domains, thereby ``translating'' an image from a domain to the other while preserving some information from the original. These methods have been applied to various applications in computer vision, such as colorization \cite{color}, super-resolution \cite{srgan,esrgan}, medical imaging and photorealistic image synthesis \cite{stackgan}.

In recent years, i2i methods have achieved promising results in terms of visual quality and diversity. CycleGAN~\cite{cycleGan} learns a one-to-one mapping between domains by introducing the cycle consistency constraint that enforces the model to preserve the content information while changing the style of the image. This idea inspired many works such as \cite{MUNIT, DRIT} to disentangle the feature space into 1) a domain-specific space for the style; and 2) a shared space for the content---this allowed more diverse multimodal generation. Recent methods such as StarGAN~\cite{stargan,starganv} unified the process in a single framework to achieve compelling results both in terms of visual quality and diversity. In the problem of reference-guided image synthesis, such methods are capable of extracting the style of a target image and merging it with the content of a source image to generate an image that shares information from both images. An inherent assumption of these methods is that both style and content images share the same spatial resolution. 

% However, most of the previous methods were only applied on high resolution (HR) images and did not consider the scenario of having a very low resolution target image. 

In this paper, we consider the scenario where the target image is of much lower resolution than the source image.
In this case, the target image only contains low frequency information such as the general shape, pose, and color of the subject.
% Since the low resolution (LR) target image lacks most of the fine details, it mostly contains low frequency information such as colors but it also contain information about the pose or general shape of objects. 
%
We therefore aim to learn a mapping that can leverage the high frequency information present in the HR source image, and combine it with low frequency information preserved in the LR target. As illustrated in fig.~\ref{fig:first}, such a mapping can learn to generate an HR image that shares distinctive features from both HR and LR inputs in an unsupervised manner.

This scenario also bears resemblance to the image super-resolution problem, which aims at generating an HR image that corresponds to a LR input. Both scenarios are highly ill-posed since there exists a large number of HR images which all downscale to exactly the same LR image---number which grows exponentially with each downscaling factor~\cite{limitsSr}. We also demonstrate that our framework can be applied to super-resolve a very low resolution image given a high resolution exemplar as guidance. 
%For example, PULSE~\cite{pulse} searches the latent space of a pretrained StyleGAN~\cite{StyleGan} for an HR image which, when downscaled, corresponds exactly to the input LR image. 
% A crucial difference here is that we consider image-to-image translation \emph{conditioned} by a low-resolution image. Such approach goes beyond the signal enhancing approach of classical super-resolution, with our proposed conditional image-to-image translation method to generate plausible versions of HR images that downscale to the same LR image. 
%This could be considered a case of ``guided'' (where the guide is the content image in HR), as opposed to ``blind'' super-resolution.
%\todo{Fig.~\ref{fig:first} should be referred to somewhere., Done !}

% Since the each LR image can correspond to multiple HR images, and with each down scaling factor that number grows exponentially \cite{limitsSr,Pulse}, which means that LR target can have an infinite  set of potential HR images. Therefore, given the LR target and the HR source image, our method learns to generate an HR image that can be considered as the high resolution version of LR target and at the same time it is directly related to the HR source image. 

% To the best of our knowledge, we are the first work to propose a solution to such task. 
The main contribution of this paper is to define a novel framework that deals with resolution mismatch between target and source for image-to-image translation. 
% After learning, the mapping can be applied in a single feedforward pass through the network. 
We demonstrate that the approach can effectively be used when dealing with very low resolution targets where details are completely blurred to the point of being visually unrecognizable. 
When evaluating our approach on the CelebA-HQ and AFHQ datasets, we show that our framework results in more realistic samples than state-of-the-art image-to-image translation methods such as Stargan-v2~\cite{starganv}.  We validated our findings by reporting FID and LPIPS scores on both dataset, and also provide more evidence by reporting density/coverage metrics~\cite{divCov}.
These extensive experiments demonstrate that our method can generate results that are photo-realistic, and that convincingly fuses information from both the HR source and LR target images. 
%An ablation study on the resolution of the target image is also provided and we finally investigated in sec~\ref{sec:ablation}, the effect of varying the resolution of the target image on the generation process. 
% We also make a qualitative comparison of the super-resolution results of our method with PULSE in sec.~\ref{sec:superres}.

\begin{figure*}[ht]
    \centering
    \subfloat[HR subspaces $\mathcal{M}_\mathbf{X}$ and $\mathcal{M}_\mathbf{Y}$ and their corresponding LR subspaces $\mathcal{R}_\mathbf{x}$ and $\mathcal{R}_\mathbf{y}$, related by a downscaling function $\mathrm{DS}(\cdot)$.]
    {\includegraphics[width=0.30 \textwidth]{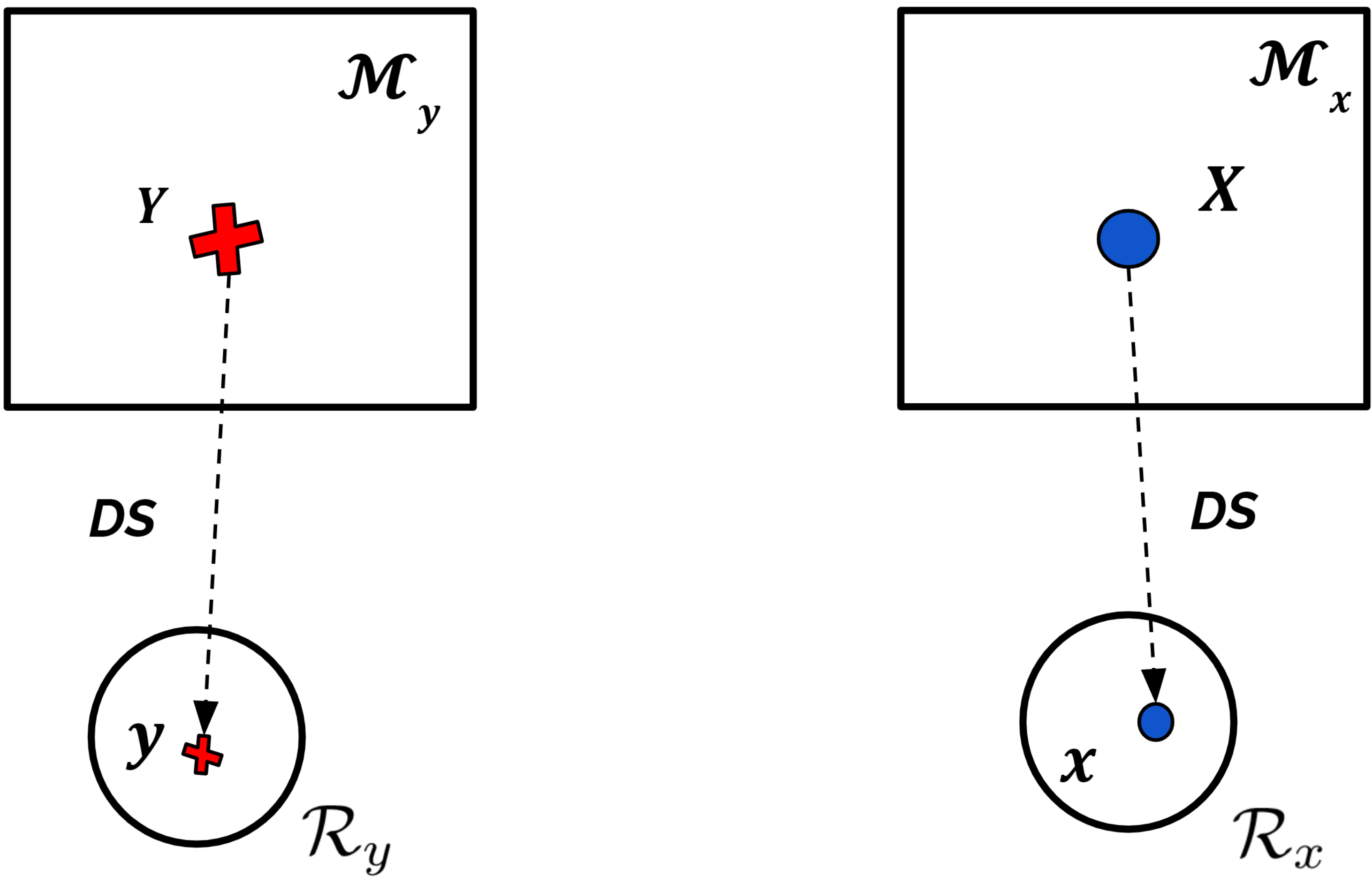}}
    \hspace{0.4cm}
    % \rulesep
    % \hspace{0.2cm}
    \subfloat[Translation of HR source image $\mathbf{Y}$ from $\mathcal{M}_\mathbf{Y}$ to $\mathcal{M}_\mathbf{X}$ guided by LR target $\mathbf{x}$.]
    {\includegraphics[width=0.30\textwidth]{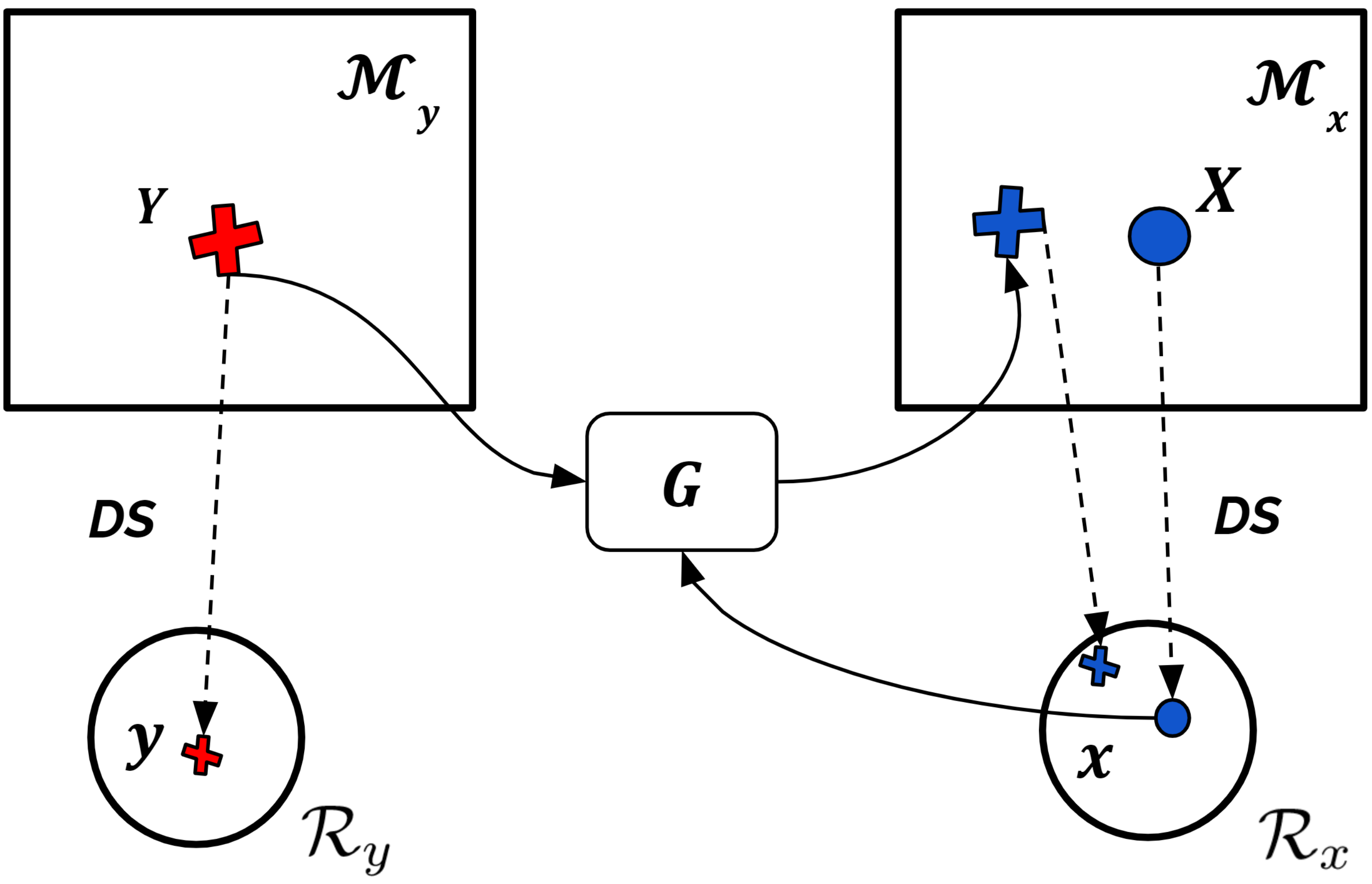}}
    \hspace{0.4cm}
    % \rulesep
    % \hspace{0.2cm}
    \subfloat[Translation of HR source image $\mathbf{X}$ from $\mathcal{M}_\mathbf{X}$ to $\mathcal{M}_\mathbf{Y}$ guided by LR target $\mathbf{y}$.]
    {\includegraphics[width=0.30\textwidth]{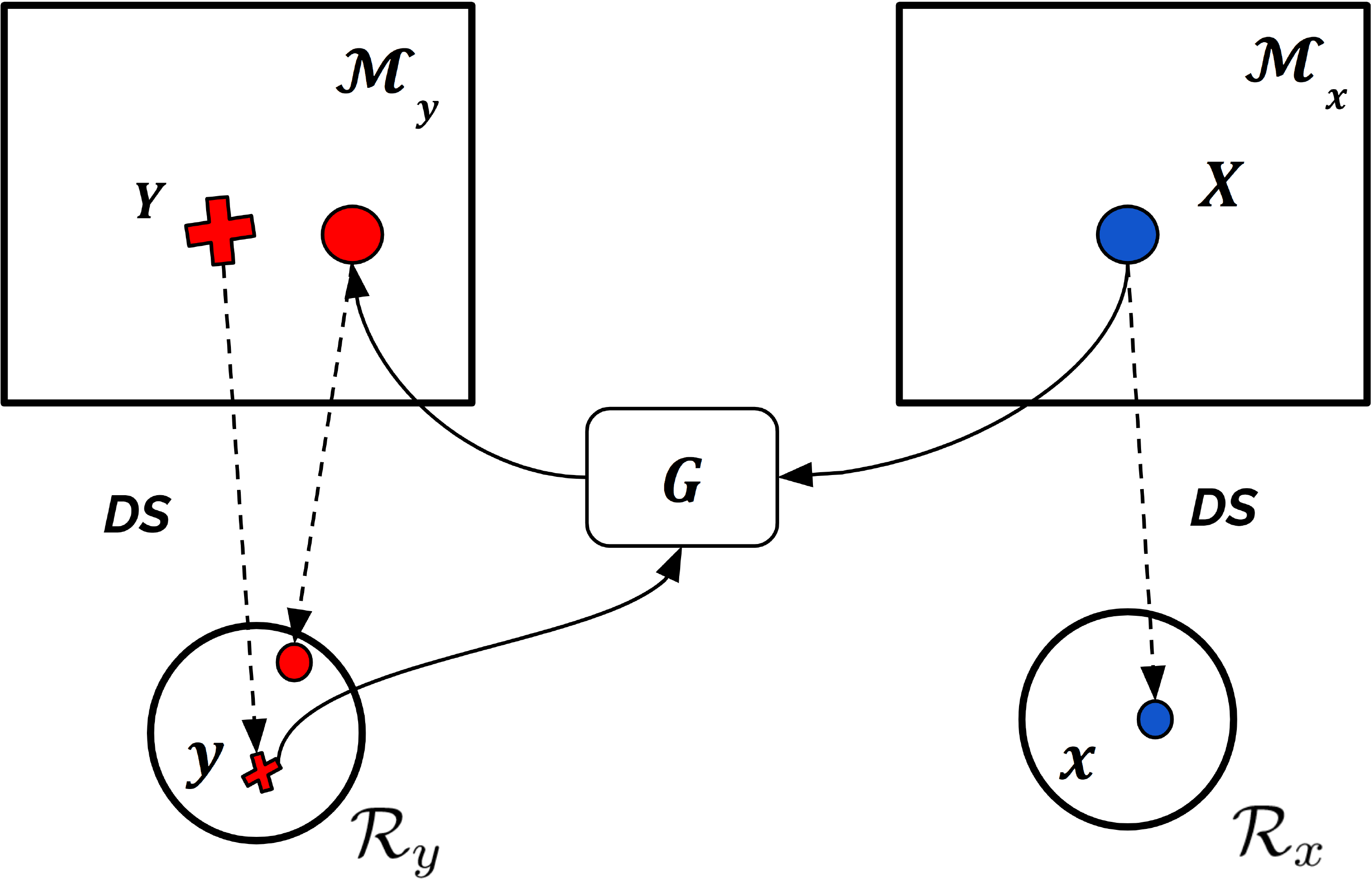}}
    \caption{Conceptual illustration of the proposed framework: the generator $G$ is trained to map a HR image $\mathbf{Y}$ (resp. $\mathbf{X}$) from its domain $\mathcal{M}_\mathbf{Y}$ (resp. $\mathcal{M}_\mathbf{X}$) to the other domain $\mathcal{M}_\mathbf{X}$ (resp. $\mathcal{M}_\mathbf{Y}$), conditioned on a LR version $\mathbf{x}$ of $\mathbf{X}$ (resp. $\mathbf{y}$ of $\mathbf{Y}$). }%
    \label{fig:illustration}%
\end{figure*}

%!TEX root = egbib.tex
\section{Related work}

Generative adversarial networks (GANs)~\cite{goodfellow} have demonstrated promising results in various applications in computer vision, including image generation~\cite{Biggan,StyleGan,stylegan2}, super-resolution~\cite{srgan,esrgan} and image-to-image translation~\cite{MUNIT,pix2pix,DRIT,msgan,cycleGan}. 

Recent work has striven to improve sample quality and diversity, notably through theoretical breakthroughs in terms of defining loss functions which provide more stable training~\cite{Wgan,wgangp,lsgan,EnegryGan} and encourage diversity in the generated images~\cite{Greg,diverseGan}. Architectural innovations also played a crucial role in these advancements~\cite{Biggan,starganv,ProGanCelebA,sagan}. For instance, \cite{sagan} makes use of an attention layer that allows it to focus on long-range dependencies present in the image. Spectral normalization~\cite{miyato2018spectral} stabilizes the network, which also translates into having high quality samples. Since our work sits at the intersection of reference-guided image synthesis and super-resolution, the remainder of this section focuses on these two branches exclusively.

%Also, recent methods~\cite{StyleGan,starganv} introduced a mapping network designed to take a normalized latent vector and output a disentangled latent vector with same size which fed to into intermediate adaptive instance normalization~\cite{Adain} layers of the generator borrowed from the style transfer literature. 

\paragraph{Reference-guided image synthesis} 
Also dubbed ``conditional''~\cite{CItoI}, reference-guided i2i translation methods seek to learn a mapping from a source to a target domain while being conditioned on a specific image instance belonging to the target domain. In this case, some methods attempt to preserve the ``content'' of the source image (identity, pose) and apply the ``style'' (hair/skin color) of the target. Inspired by the mapping network of StyleGAN~\cite{StyleGan}, recent methods~\cite{stargan,starganv,MUNIT} make use of a style encoder to extract the style of a target image and feed it to the generator, typically via adaptive instance normalization (AdaIN)~\cite{Adain}. All these methods have the built-in assumption that the resolution of the source and target images are the same. In this work, we explore the scenario where the target (style) image has much lower resolution than the source. 

% Although these methods are conditioned solely on target images from the designated domain with the same resolution, the proposed technique is conditioned on a very low resolution version of the target image.

\paragraph{Super-resolution}
Our approach is also related to super-resolution methods, which are aiming to learn a mapping from LR to HR images~\cite{srgan,dong2014learning}. Other approaches leverage knowledge about the class of images for super-resolution. For example, when dealing with faces, explicit facial priors can be leveraged~\cite{chen2018fsrnet}. Of particular interest, PULSE~\cite{pulse} leverages a pretrained StyleGAN~\cite{StyleGan} and, through an iterative optimization procedure, searches the space of latent style vectors for a face which downscales correctly to a given LR image. Our method bears resemblance to PULSE, but allows for a guided and therefore more controlled generation procedure with a single forward pass in the network. 

More closely related are the so-called reference-guided super-resolution methods, which, in addition to the LR input image, also accept additional HR images for guidance. Here, the reference images need to contain similar content (e.g.\ textures) as the LR image. Representative recent methods propose to transfer information using cross-scale warping layers~\cite{CrossCSR} or with texture transfer~\cite{srntt}. In contrast, our method frames super-resolution in an i2i context, relying on a specific instance (e.g.\ a specific person for faces) to guide the super-resolution process.

\section{Method}
\subsection{Formulation}

Given a LR target image $\mathbf{x} \in \mathbb{R}^{m \times n}$, we define the associated $\mathcal{R}_\mathbf{x}$ subspace of LR images as:
\begin{equation}
\mathcal{R}_\mathbf{x}=\left\{\forall i, \mathbf{x}_i \in \mathbb{R}^{m \times n}:\left\|\mathbf{x}_i - \mathbf{x}\right\|_{p} \leq \epsilon\right\}  \,.
\label{Rx}
\end{equation}
We consider an $\epsilon$ close to zero, so that each subspace contains only the LR images $\mathbf{x}_i$ that are highly similar  to the LR target $\mathbf{x}$ according to a given norm $p$ (here $p=1$).
We are also defining a subspace $\mathcal{M}_\mathbf{X}$ in the HR image manifold that includes all HR images that are included in $\mathcal{R}_\mathbf{x}$ when downscaled as LR images:
\begin{equation}
\mathcal{M}_\mathbf{X} = \left\{\forall j, \mathbf{X}_j \in \mathbb{R}^{M \times N}: \mathrm{DS}(\mathbf{X}_j) \in \mathcal{R}_\mathbf{x} \right\}   \,,
\end{equation}
where $\mathbf{X}_j$ is an image in the HR space, $\mathbf{X}\in \mathbb{R}^{M \times N}$ is the one in the HR space corresponding to $\mathbf{x}$, and $\mathrm{DS}(\cdot)$ is a downscaling procedure. Therefore, for each subspace $\mathcal{M}_\mathbf{X}$ on the HR image manifold, there exists a corresponding subspace $\mathcal{R}_\mathbf{x}$ in the LR space related by  $\mathrm{DS}(\cdot)$. 
As illustrated in fig.~\ref{fig:illustration}, given two HR-LR image pairs $(\mathbf{X} \in \mathcal{M}_\mathbf{X},\,\mathbf{x} \in \mathcal{R}_\mathbf{x})$ and $(\mathbf{Y} \in \mathcal{M}_\mathbf{Y},\,\mathbf{y} \in \mathcal{R}_\mathbf{y})$, our goal is to learn a function $G$ that translates the HR images from one subspace $\mathcal{M}_\mathbf{X}$ to another subspace $\mathcal{M}_\mathbf{Y}$. These two subspaces are identified by their LR counterpart (i.e., $\mathcal{R}_\mathbf{x}$ and $\mathcal{R}_\mathbf{y}$ respectively), using LR images as additional information to specify the target HR space: 
\begin{equation}
    % \left\{
    \begin{split}
    G:& \, \mathcal{M}_\mathbf{Y} \times \mathcal{R}_\mathbf{x} \mapsto \mathcal{M}_\mathbf{X}, \; G(\mathbf{Y},\mathbf{x};\theta)\,, \\
    G:& \, \mathcal{M}_\mathbf{X} \times \mathcal{R}_\mathbf{y} \mapsto \mathcal{M}_\mathbf{Y}, \; G(\mathbf{X},\mathbf{y};\theta)
    \,.
    \end{split} 
    % \right\}
\end{equation}
\begin{figure}[t]
\centering
%\fbox{\rule{0pt}{2in} \rule{.9\linewidth}{0pt}}
\includegraphics[width=\linewidth]{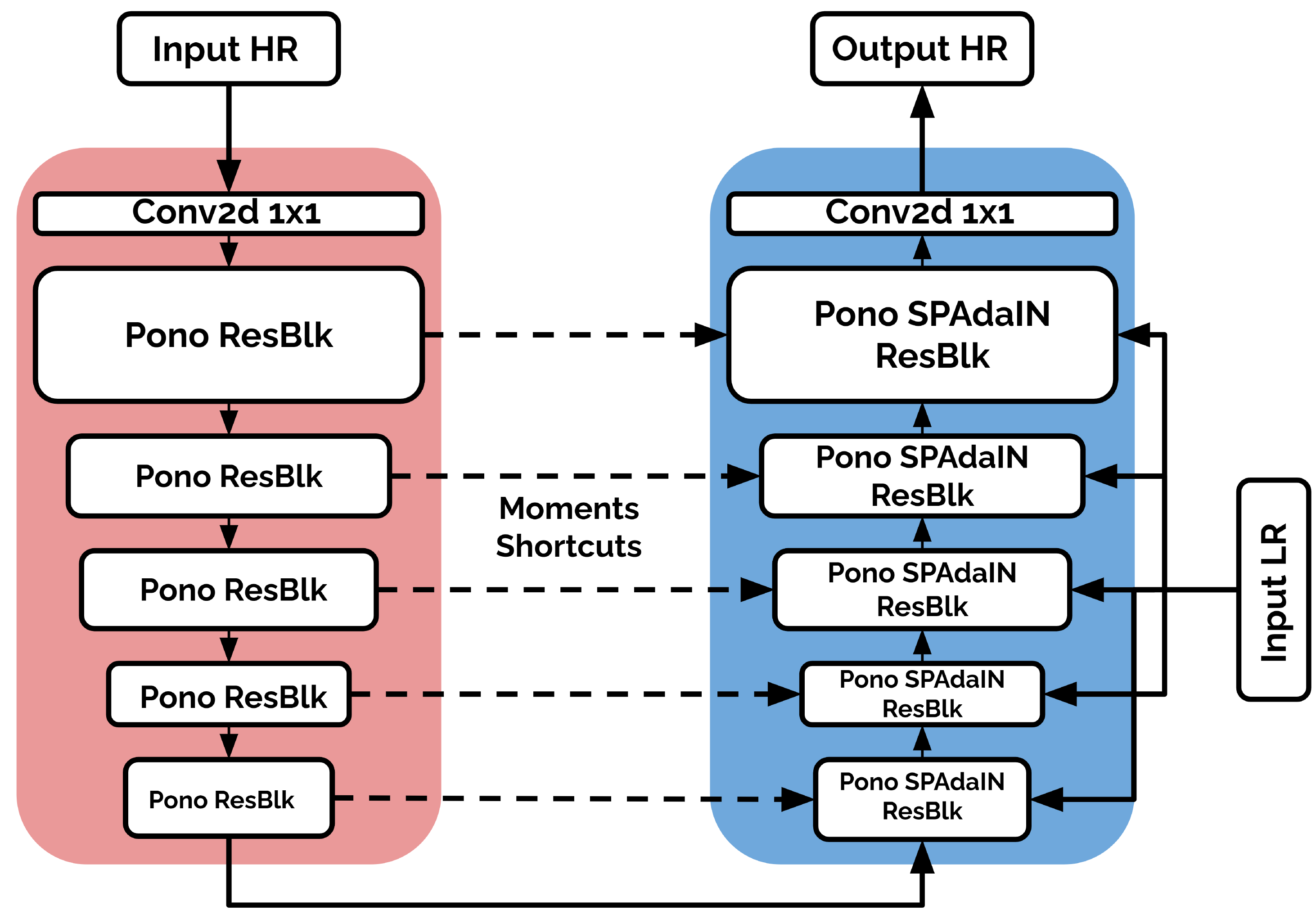} 
\caption{Generator architecture: the encoding part (in red) is used for extracting structural information in the source image, which is merged in the decoder (in blue) with the incoming features from the decoding part with LR targets. }
\label{fig:generator}
\end{figure}
Following conventional GAN terminology~\cite{goodfellow}, parameterized function $G(\cdot,\cdot;\theta)$ is a generator (simplified as $G(\cdot,\cdot)$ hereinafter). The goal of $G$ is to translate a HR image from a HR subspace to another HR subspace while preserving some of the original image information (i.e., high-frequency content). For training $G$, we are using a discriminator $D$ that plays two roles: 1) to classify whether the generated images are fake or real, pushing $G$ to sample from the natural image manifold; and 2) to judge whether the translated high resolution image is part of the right LR subspace when downscaled, guiding $G$ into the correct subspace. 
% Thus our method is different form the traditional image-to-image translation method, since the domains are subspaces of the manifold identifiable by the LR target and not prefixed. 

\subsection{Training Objectives}

For achieving these two discrimination roles, the GAN is trained according to the following minimax optimization:
\begin{equation}
\min_{G} \max_{D} ~ \mathcal{L}_\mathrm{adv} + \lambda_\mathrm{cyc} \mathcal{L}_\mathrm{cyc} \,,
\label{eq:fullobj}
\end{equation}
where $\mathcal{L}_\mathrm{adv}$ is an adversarial loss used to ensure that we are generating plausible natural images, $\mathcal{L}_\mathrm{cyc}$ is the cycle consistency loss that is getting the translated images are kept on the correct subspace while carrying the right information, and $\lambda_\mathrm{cyc}$ is a hyperparameter to achieve the right balance between these two losses.

At each training iteration, four forward passes are done with the generator, the first two for translating from a subspace to the other one in both ways, $G(\mathbf{X},\mathbf{y})$ and $G(\mathbf{Y},\mathbf{x})$, while the other two forward passes are for the cycle consistency constraint, $G(G(\mathbf{X},\mathbf{y}),\mathbf{x})$ and $G(G(\mathbf{Y},\mathbf{x}),\mathbf{y})$. The discriminator is used to make sure that the generated samples are from the designated subspace.

\paragraph{Adversarial loss} 
Following \cite{Lag}, we provide the discriminator with the absolute difference $\mathbf{d}_{\mathbf{X}\rightarrow \mathcal{M}_\mathbf{Y}}\in\mathbb{R}^{m\times n}$ between the downscaled version of the generated image $\mathrm{DS}(G(\mathbf{X},\mathbf{y}))$ and the LR target $\mathbf{y}$:
\begin{equation}
    \mathbf{d}_{\mathbf{X}\rightarrow \mathcal{M}_\mathbf{Y}} = \frac{\left|\mathbf{y}-\left\lfloor r\, \mathrm{DS}(G(\mathbf{X}, \mathbf{y}))\right\rfloor\right|}{r}\,,
    \label{eq:ste}
\end{equation}
where $r$ is the color resolution. As in \cite{Lag}, we round the downscaled image to its nearest color resolution ($r=2/255$, since pixel values are in $[-1,1]$) to avoid unstable optimization caused by exceedingly large weights to measure small pixel value differences. A \emph{straight through} estimator \cite{steEstimator} is employed to pass the gradient through the rounding operation in eq.~\ref{eq:ste}. The discriminator therefore takes as inputs:
\begin{equation}
    \left\{
    \begin{array}{ll}
    D(\mathbf{Y}, \mathbf{0}) & \text{for real samples,} \\
    D\left(G(\mathbf{X}, \mathbf{y}),  \mathbf{d}_{\mathbf{X}\rightarrow \mathcal{M}_\mathbf{Y}} \right) & \text {otherwise.}
    \end{array} \right.
    \label{eq:disInput}
\end{equation}
Here, $\mathbf{0}$ is an all-zeros $m\times n$ image difference, since the downscaled version of $\mathbf{Y}$ is exactly $\mathbf{y}$. However, for fake samples, the absolute difference $\mathbf{d}_{\mathbf{X}\rightarrow \mathcal{M}_\mathbf{Y}}$ depends on how close is the generator to the designated subspace, $\mathcal{M}_\mathbf{Y}$ in our example. Both networks $G$ and $D$ are trained via the resulting adversarial loss:
\begin{align}
  \mathcal{L}_\mathrm{adv} = & ~ \mathbb{E}_\mathbf{X}\left[\log D(\mathbf{X}, \mathbf{0})\right] + \mathbb{E}_\mathbf{Y}\left[\log D(\mathbf{Y}, \mathbf{0})\right]\nonumber\\
  & + \mathbb{E}_{\mathbf{X}, \mathbf{y}}\left[\log \left(1-D(G(\mathbf{X}, \mathbf{y}),\mathbf{d}_{\mathbf{X}\rightarrow \mathcal{M}_\mathbf{Y}} )\right)\right]\,.
\end{align}

\paragraph{Cycle consistency loss}  To make sure that the generator $G$ preserves the high frequency information available in the source HR image, we employ the cycle consistency constraint~\cite{pix2pix,Unit,cycleGan} in both directions, each time by changing the LR target to specify the designated subspace: 
\begin{align}
    \mathcal{L}_\mathrm{cyc} = & ~ \mathbb{E}_{\mathbf{X}, \mathbf{x}, \mathbf{y}} \|\mathbf{X}-G(G(\mathbf{X},\mathbf{y}),\mathbf{x})\|_1\nonumber\\
    & + \mathbb{E}_{\mathbf{Y}, \mathbf{x}, \mathbf{y}}\|\mathbf{Y}-G(G(\mathbf{Y},\mathbf{x}),\mathbf{y})\|_1\,.
\end{align}
This cycle consistency loss encourages the generator to identify for the shared/invariant information between each two subspaces and preserve it during translation.

\subsection{Architecture}
\begin{figure*}[ht]
    \centering
    \subfloat[Pono residual blocks]{\includegraphics[width=0.42 \textwidth]{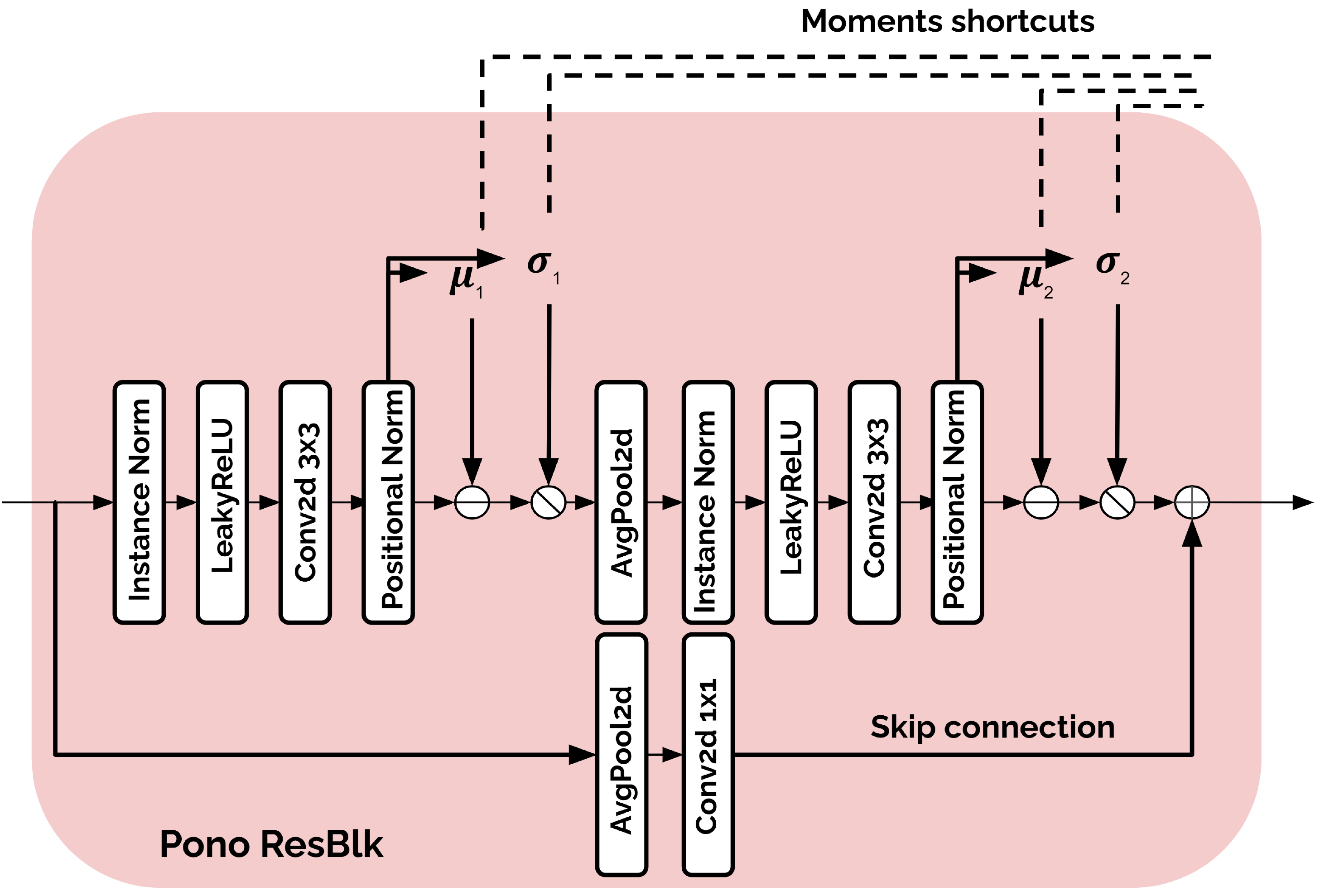}
    \label{fig:res a}}%
    \qquad
    \subfloat[Pono SPAdaIn residual blocks]{\includegraphics[width=0.42\textwidth]{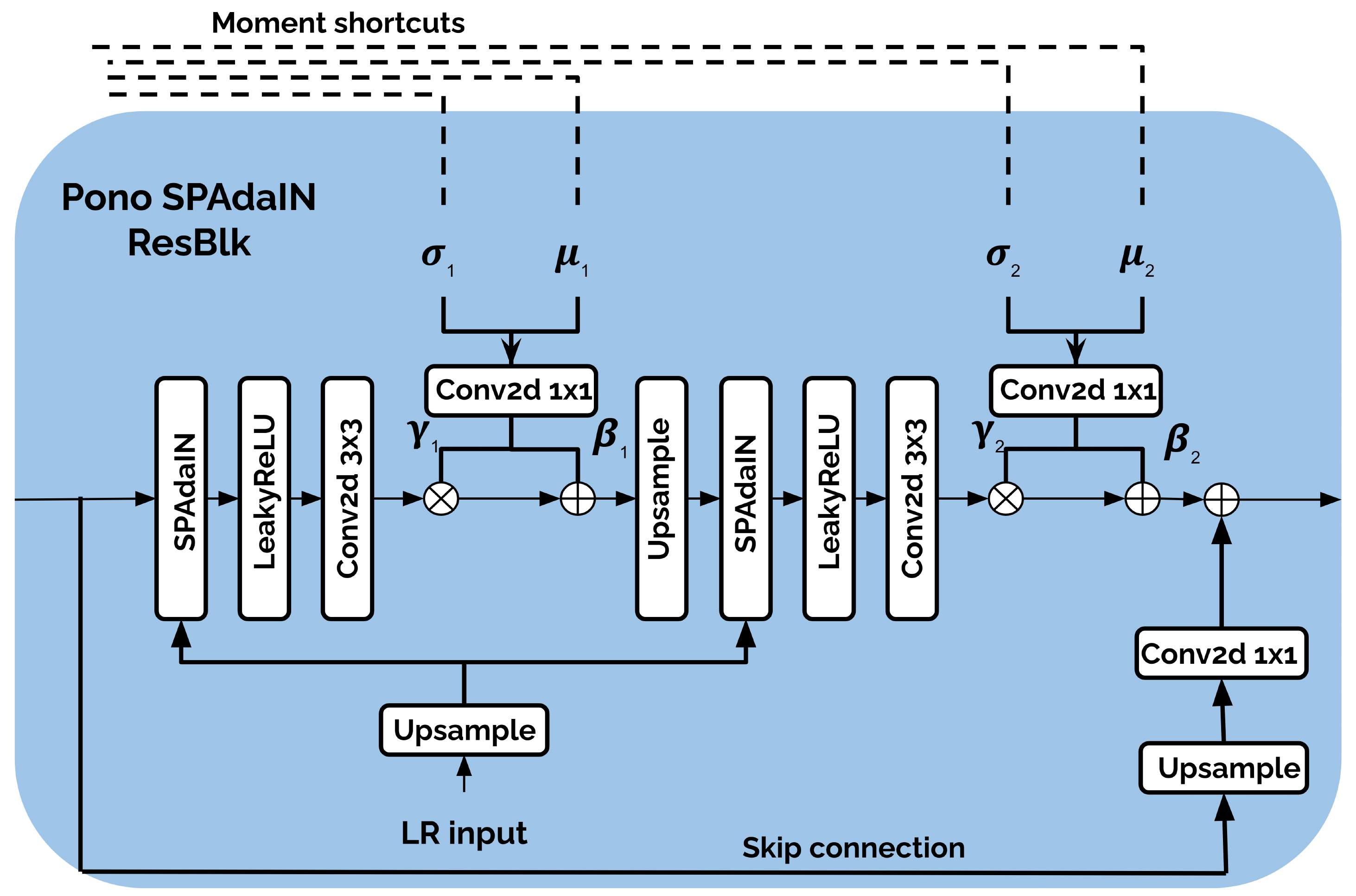}
    \label{fig:res b}}%
    \caption{Residual blocks used in the generator $G$: (a) Encoding pono residual blocks, used to extract and compress the features coming from the source image and pass it to the decoding part of the generator through moments shortcuts and as compressed features  ; and (b) Decoding pono SPAdaIn residual blocks takes as input extracted features from the encoding part, and also upsample the LR target image and pass it to the SPAdaIn layer \cite{NeuralPose} }%
    \label{fig:resblocks}%
\end{figure*}

Most of the recent image-to-image translation models (e.g.~\cite{stargan,starganv,MUNIT}) rely on Adaptive Instance Normalization (AdaIN)~\cite{Adain, StyleGan} to transfer the style from a reference image to a source image. 
In our work, however, the hypothesis of content and style is not suitable since the LR image contains information on both style (e.g., colors) and content (e.g., pose). Thus, our generator adapts HR source image to the content and style of the LR image through the use of spatially adaptive instance normalization (SPAdaIN)~\cite{NeuralPose}.
% Therefore, the use of AdaIN~\cite{Adain} in our generator does not server our need, since, it s shown in \cite{NeuralPose} that AdaIN, is not suitable for shapes and pose transfer, due to the lack of learnable parameters, which will only push the network to adopt the present shapes and pose of the input high resolution image.

% Since our task, requires that the generated images when down-scaled is spatially aligned with the corresponding LR of the targeted subspaces. 

Generator $G$ (fig.~\ref{fig:generator}) is U-shaped, with skip connections (moments shortcuts \cite{Pono}) between the encoding and decoding part. The encoder takes the input HR image, passes it through a series of downsampling residual blocks (ResBlks)~\cite{he2016deep}. Each ResBlks is equipped with instance normalization (IN) to remove the style of the input, followed by 2D convolution layers and a positional normalization (Pono)~\cite{Pono}. The mean $\mu$ and variance $\sigma$ are subtracted and passed as a skip connection to the corresponding block in the decoder. Pono and moments shortcuts plays a crucial role in transferring the needed structural information from the HR to the decoding part of the network. These blocks, dubbed \emph{Pono ResBlks}, are illustrated in detail in fig.~\ref{fig:res a}.

For the decoder blocks (fig.~\ref{fig:res b}), we use SPAdaIN~\cite{NeuralPose} conditioned on the LR image, where the LR image is first upsampled to the corresponding resolution of the \emph{Pono SPAdaIN ResBlk} using bilinear upsampling. It is then followed by 2D convolution layers and a dynamic moment shortcut layer, where, instead of reinjecting $\mu$ and $\sigma$ as is, we use a convolutional layer that takes $\mu$ and $\sigma$ as inputs to generate the $\beta$ and $\gamma$ used as moment shortcuts. Using the dynamic version of moment shortcuts allows the network to adapt and align the shape of the incoming structural information to its LR counterpart.

We use the StarGAN~v2~\cite{starganv} discriminator architecture minus the domain-specific layers since we do not have predefined domains. We also concatenate the image difference  (eq.~\ref{eq:ste}) at the corresponding layer (same height and width).
%!TEX root = egbib.tex
\section{Experiments}

\paragraph{Baseline} We compare our method with Stargan-v2~\cite{starganv}, the state-of-the-art for image-to-image generation on CelebA-HQ and AFHQ.

\paragraph{Datasets} We evaluate our method on the CelebA-HQ~\cite{ProGanCelebA} and AFHQ~\cite{starganv} datasets. However, for CelebA-HQ we do not separate the two domains into female and male, since both domains are close to each other. Also, we are not using any extra information (e.g. facial attributes of CelebA-HQ). As for AFHQ, we train our network on each domain separately, since the amount of information shared between these is much lower. Average pooling is used as downscaling operator to generate the LR images, as in \cite{Lag}.

\paragraph{Evaluation metrics} Baseline results are evaluated according to the metrics of image-to-image translation used in \cite{MUNIT, DRIT, msgan}. Specifically, diversity and visual quality of samples produced by different methods are evaluated both with the Fréchet inception distance (FID)~\cite{ttur} and the learned perceptual image patch similarity (LPIPS)~\cite{LPIPS}. Since the FID score entwine diversity and fidelity in a single metric~\cite{divCov,precRecal}, we also experiment with the density and coverage metrics proposed in \cite{divCov}.

%-------------------------------------------------------------------------
\subsection{Training Setup}

For our experiments, we fixed the LR image resolution to $8\times8$ and experimented with $128 \times 128$ and $256 \times 256$ for the HR image resolution---we ablate the effect of LR image resolution in sec.~\ref{sec:ablation}. 
We train our networks with Adam~\cite{adam} and TTUR~\cite{ttur}, with a learning rate of $10^{-3}$ for the generator and $4 \times 10^{-3}$ for the discriminator. We also used $R^1$ regularization~\cite{Greg} with $\gamma=0.5$, with a batch size of 8. Spectral normalization~\cite{spectralNorm} was used in all the layers of both $G$ and $D$. In eq.~\ref{Rx}, we use $\epsilon=0$ to push the downscaled version of the generated image to be as close as possible to the LR target. We set $\lambda_{cyc}=1$ when trained on $128 \times 128$, and to $\lambda_{cyc}=0.1$ for $256\times256$.

    \begin{figure*}
    \begin{center}
    
    \includegraphics[width=0.9\linewidth]{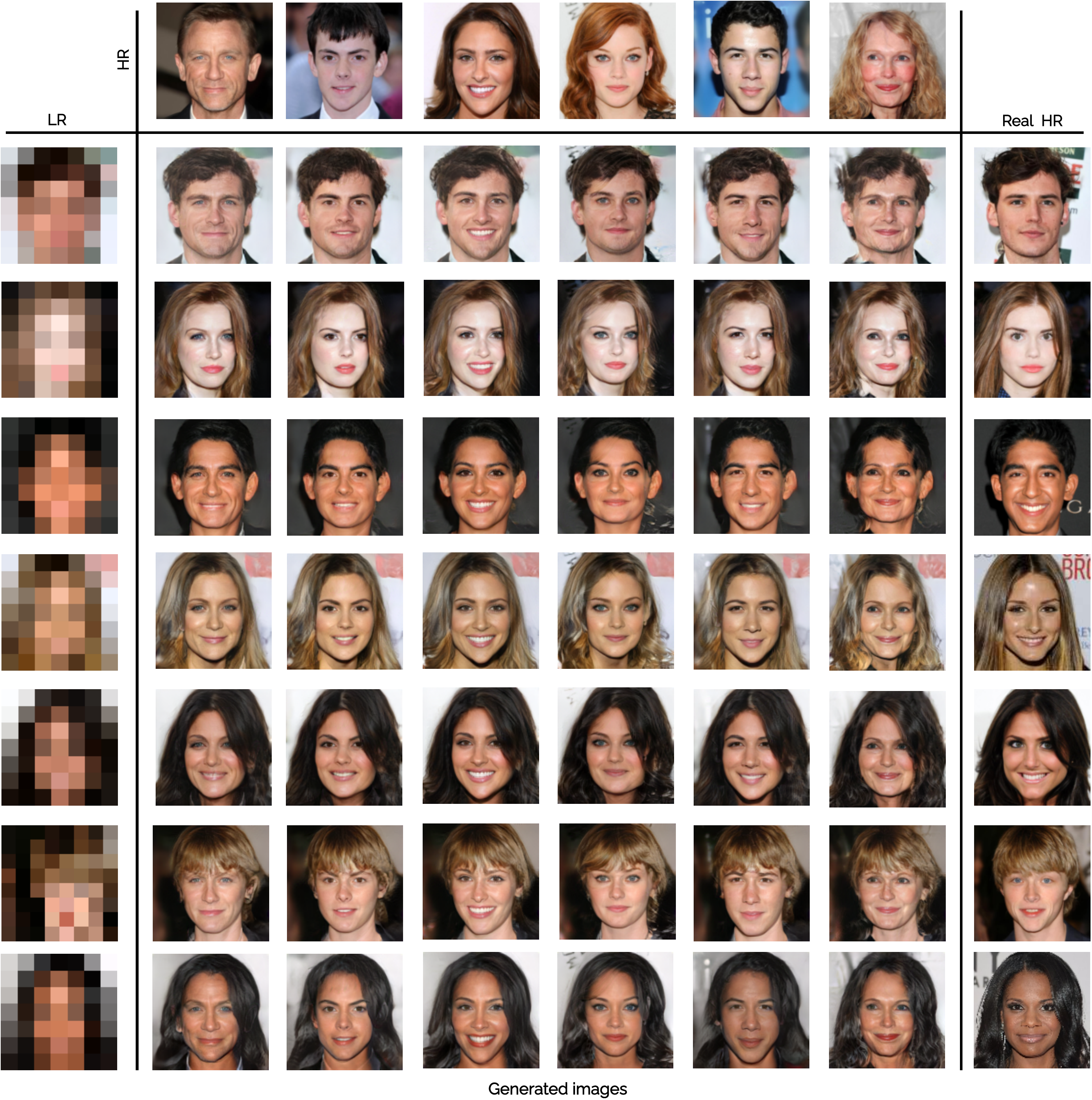}
    \end{center}
       \caption{Qualitative reference-guided image synthesis results on CelebA-HQ. Our method takes the HR source images (top row), and translates them according to the LR target (left column). We also add the real HR target (not seen by the network) for visual comparison. \textbf{See supplementary material for more results}.}
    \label{fig:lots}
    \end{figure*}

\begin{figure*}
    \begin{center}
    \hspace{-1cm}
    \includegraphics[width=0.88\linewidth]{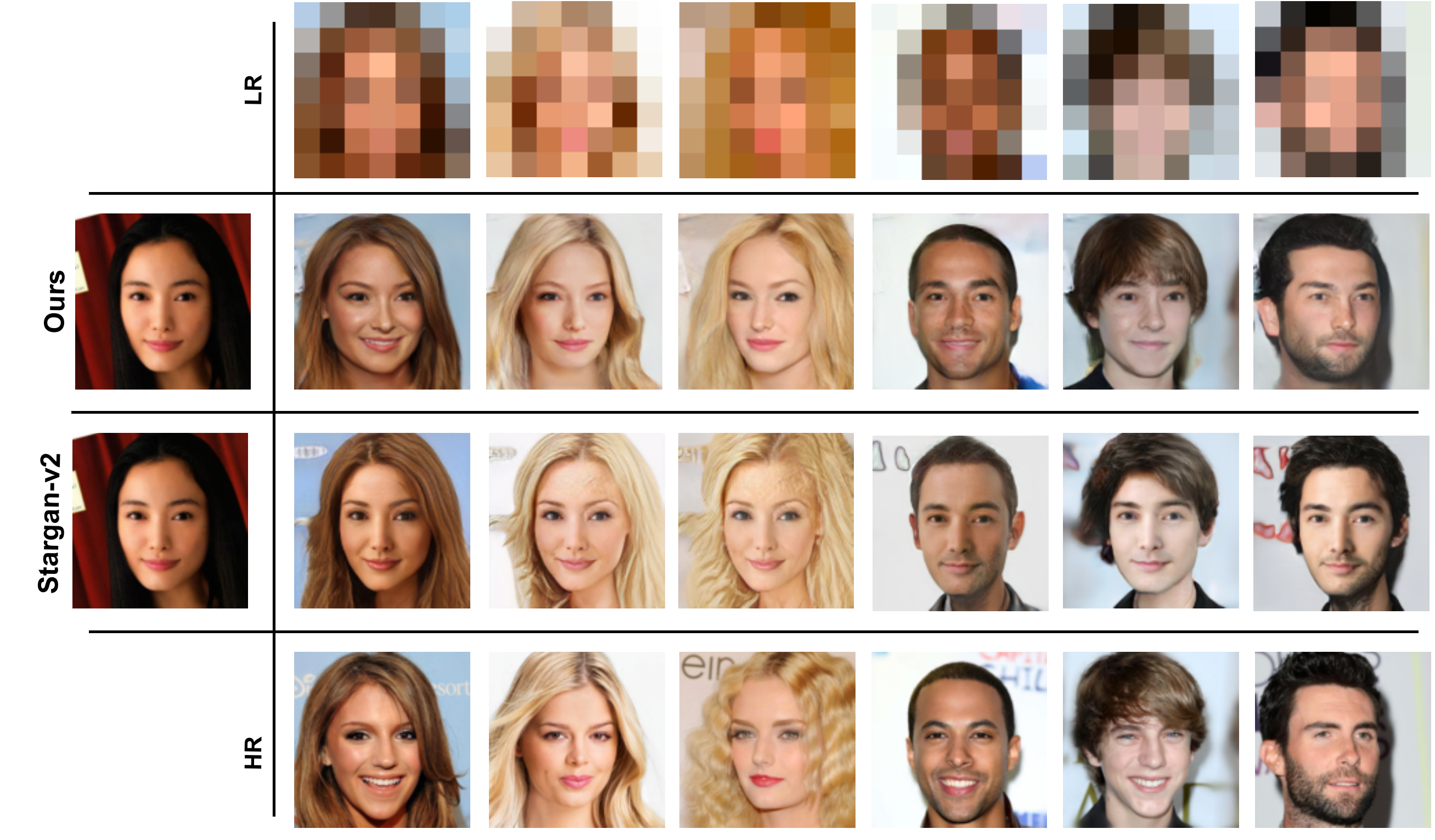}
    \end{center}
       \caption{Comparison between our method and reference guided Stargan-v2~\cite{starganv} on CelebA-HQ. For both methods, we use the same HR image as source (left column). For the reference image, our method uses the LR images (top row) while Stargan-v2~\cite{starganv} uses the corresponding HR (bottom row).}
    \label{fig:comparison}
    \end{figure*}

\begin{figure}
    \begin{center}
    %\hspace{-1cm}
    \includegraphics[width=\linewidth]{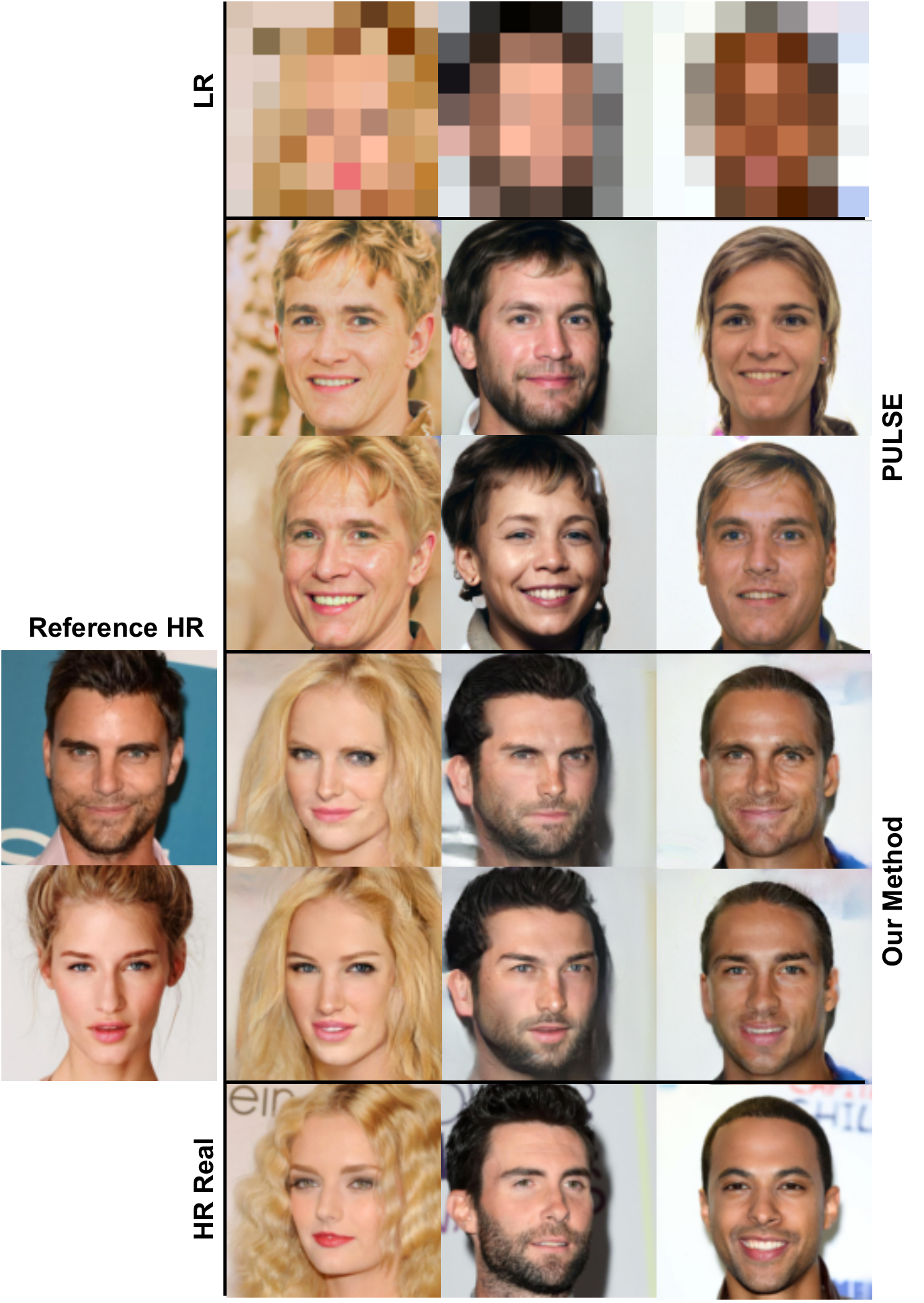}
    \end{center}
       \caption{Comparison between our method and PULSE~\cite{pulse} on CelebA-HQ. While both methods use the same LR image (top row), ours also leverages the reference images (left column) to guide the generation process. We also show the corresponding HR (bottom row) for reference only---neither method sees this HR image.}
    \label{fig:pulse}
    \end{figure}
%-------------------------------------------------------------------------
\subsection{Qualitative Evaluation}

Fig.~\ref{fig:comparison} compares images obtained with our framework with those obtained with Stargan-v2~\cite{starganv} using reference-guided synthesis on CelebA-HQ. Since our method focuses on generating images that downscale to the given LR image, the generator learns to merge the high frequency information presents in the source image with the low frequency information of the LR target, while preserving the identity of the person and other distinctive features. Differently from traditional i2i methods that only change the style of the source image while preserving its content, our method adapts the source image to the pose of the LR target. More qualitative samples obtained with our technique are shown in fig.~\ref{fig:lots}, where the first row of HR images are used as source images and the first column is the LR target. We also display the real HR target to show that our model is capable of generating diverse images that are different from the target.

\begin{figure}
\begin{center}

\includegraphics[width=0.98\linewidth]{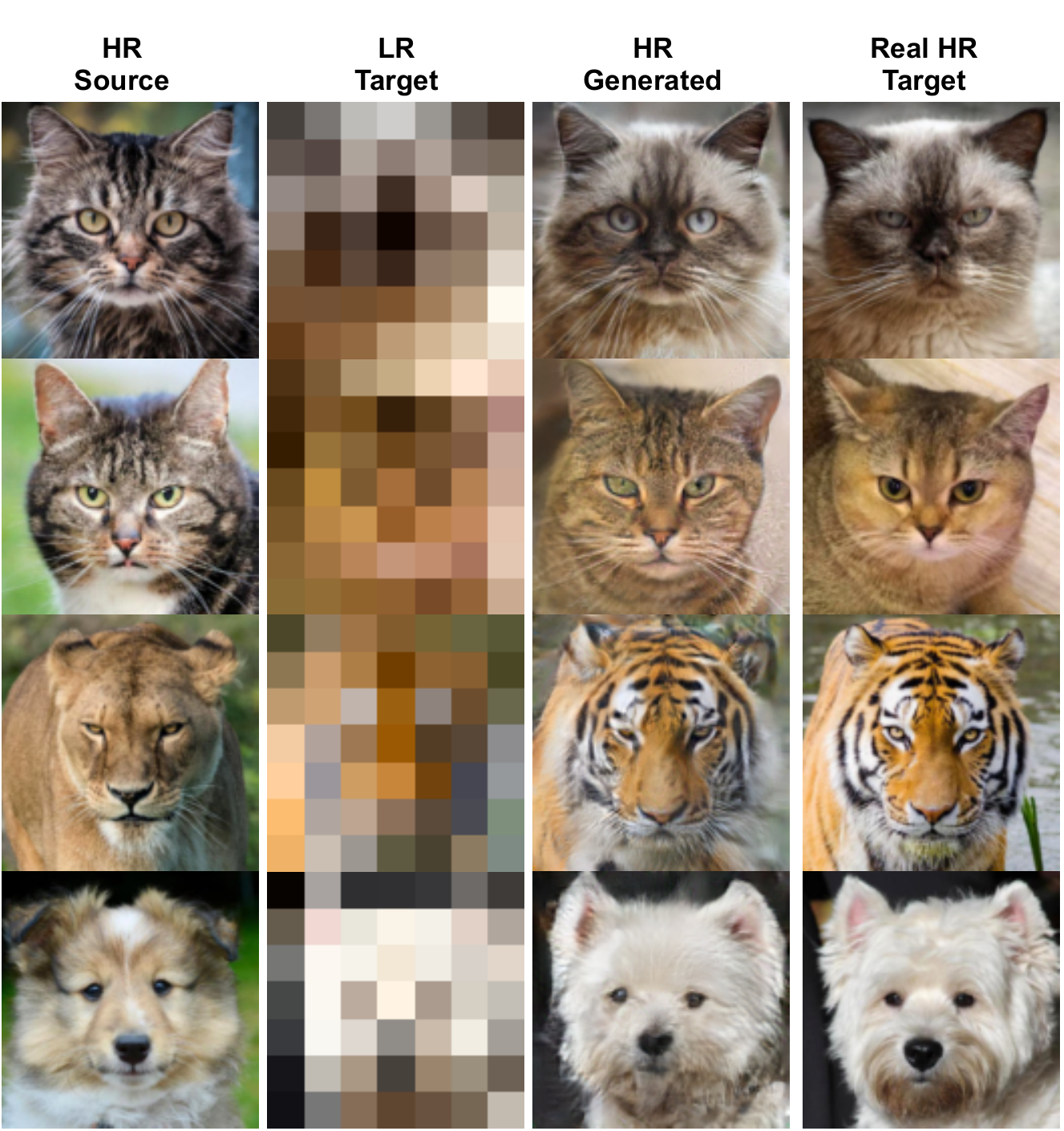}
\end{center}
   \caption{Qualitative results on the AFHQ dataset. Our method takes the HR images, and translate them to the corresponding HR subspace of the given LR target images.}
\label{fig:afhq}

\end{figure}

Fig.~\ref{fig:afhq} displays generated samples on AFHQ. Visually, we notice that our model is capable of merging most of the high frequency information coming from HR source image with low frequency information present in the LR target. The degree of this transfer depends on how much information is shared between the domain.  

%-------------------------------------------------------------------------
\subsection{Quantitative Evaluation}
\label{sec:results}
\begin{table}
\centering
\resizebox{\linewidth}{!}{
\begin{tabular}{lcccc}
\toprule
HR image res. & \multicolumn{2}{c}{$128\times 128$} & \multicolumn{2}{c}{$256\times 256$}  \\
Metric     & FID$_\downarrow$   & LPIPS$_\uparrow$           & FID$_\downarrow$        & LPIPS$_\uparrow$     \\ 
\midrule
MUNIT~\cite{MUNIT}          & --            & --              & 107.1           & 0.176      \\
DRIT~\cite{DRIT}            & --            & --              & 53.3           & 0.311      \\
MSGAN~\cite{msgan}          & --            & --              & 39.6            & 0.312      \\
Stargan-v2~\cite{starganv}  & 19.58         & 0.22            & \textbf{23.8}   & \textbf{0.38} \\
Ours                        &\textbf{15.52} & \textbf{0.34}   & 25.89           & 0.329       \\
\bottomrule

\end{tabular}}
\caption{Quantitative comparison on the CelebA dataset, comparing our method to other reference-guided i2i methods~\cite{starganv,MUNIT, DRIT, msgan}. We follow same procedure as in Stargan-v2~\cite{starganv}, but for our method we sample ten HR images for each LR image.}
\label{tab:metrics_fid}
\vspace{-0.5cm}
\end{table}

In table~\ref{tab:metrics_fid}, we report FID and LPIPS scores on the results obtained on CelebA-HQ, using two different resolutions, $128\times 128$ and $256\times 256$. Results with the $256\times 256$ resolution show a significantly lower FID of our method compared to \cite{MUNIT, DRIT, msgan}, while being similar to Stargan-v2~\cite{starganv}. We notice a better FID score with the lower $128\times 128$ resolution, being then significantly better than Stargan-v2. This is due to the fact that the task is harder with higher scale factor since we need to hallucinate more detailed textural information missing from the LR target.

For a deeper insight on the differences between our method and Stargan-v2, we used the density and coverage metrics of \cite{divCov}. The density measures the overlap between the real data and generated samples, while the coverage measures their diversity, by measuring the ratio of real samples that are covered by the fake samples~\cite{divCov}. Following \cite{divCov}, we used the feature space embedding of both the real and fakes images with a pretrained VGG16~\cite{vgg} on ImageNet. The density metric is then obtained from the $k$-nearest neighbours (with $k=5$ as in \cite{divCov}) on the 4096 features obtained from the VGG network's second fully connected layer. Diversity results reported in table~\ref{tab:diversity} show higher density values for our method, meaning that their samples are closer to the real data distribution than Stargan-v2. Higher coverage measures are also obtained for our method, meaning a better coverage of the data distribution modes. This is noticeable for the $128\times 128$ resolution, since the coverage is close to maximum value (the domain is $[0,1]$), being almost 10\% higher than Stargan-v2 while this gap doubles when the HR value is increased to $256\times 256$. This indicates that our model produces more realistic results by exploiting HR information from the source image, while being more diverse by staying faithful to the LR target image.

We also report quantitative results on AFHQ~\cite{starganv} in table~\ref{tab:afhq_metrics}, where we train our model on each domain separately given the higher differences of the domains distributions. Indeed, we found that our method excels on domains where images are structurally similar and share information, such as the ``cat'' domain. However, the ``dog'' and ``wild'' domains show a wider variety of races and species, meaning less information shared between images from the same domain. This reduces the amount of shared information that can be transferred from the HR source, forcing to hallucinate more details out of LR targets. This is confirmed by the lower LPIPS results obtained by our approach compared to Stargan-v2, while keeping similar FID scores.

\subsection{Ablation}
\label{sec:ablation}

Table~\ref{tab:resolution} illustrates the impact of LR resolution on CelebA-HQ. As the LR target resolution increases, the model exploits the information in the LR target more and more over the information provided by the HR source image. This is confirmed by sustained decrease the LPIPS score when the resolution increases from $8\times 8$ to $32\times 32$, while maintaining similar FID score. The effect of color in the LR target images is also ablated in the supp. material.

\subsection{Comparison to Super-Resolution}
\label{sec:superres}

We now compare our method to PULSE~\cite{pulse}, which super-resolves a face image by optimizing on the latent space of a pretrained StyleGAN model. Fig.~\ref{fig:pulse} presents results that illustrate similarities and differences between PULSE and our method. Both methods generate images that are realistic and faithful to the given LR image. 
From a super-resolution standpoint, our method would be considered reference-guided---but as opposed to image synthesis which is guided by the LR image, the super-resolution reference is another HR face image. We find this provides a certain amount of control over the diversity of the results, which is not possible with PULSE. Our generated samples are therefore much closer to the HR reference image.
% In addition, our generated samples are much closer to the ground truth images. This likely comes from the consistency constraint that regularizes the network and prevents from diverging too far from the real HR image.

\begin{table}
\centering
\resizebox{\linewidth}{!}{%
\begin{tabular}{lcccc}
\toprule
HR image res. & \multicolumn{2}{c}{$128\times 128$} & \multicolumn{2}{c}{$256\times 256$} \\
Metric    & Density$_\uparrow$ & Coverage$_\uparrow$ & Density$_\uparrow$ & Coverage$_\uparrow$ \\ 
\midrule    
Stargan-v2 & 1.63    & 0.89     & 1.67   & 0.69     \\
Ours       & \textbf{1.97}    & \textbf{0.98}     & \textbf{1.87}   & \textbf{0.92}     \\ 
\bottomrule
\end{tabular}%
}
\caption{Diversity and coverage metrics~\cite{divCov} comparison on CelebA-HQ with different HR resolutions.}
\label{tab:diversity}
%\vspace{-0.4cm}
\end{table}

\begin{table}
\centering
\resizebox{\linewidth}{!}{%
\begin{tabular}{lcccccc}
\toprule
Category & \multicolumn{2}{c}{Cats}     & \multicolumn{2}{c}{Dogs}             & \multicolumn{2}{c}{Wild} \\ 
Metric & FID$_\downarrow$  & LPIPS$_\uparrow$     & FID$_\downarrow$   & LPIPS$_\uparrow$     & FID$_\downarrow$         & LPIPS$_\uparrow$       \\ 
\midrule
Stargan-v2  & 25.2 & \textbf{0.42} & \textbf{56.5} & 0.34 & \textbf{19.87} & \textbf{0.46} \\
Ours        & \textbf{22.8} & 0.40      & 67.3 & \textbf{0.47}      & 20.61       & 0.23        \\ 
\bottomrule
\end{tabular}%
}
\caption{Quantitative comparison on the AFHQ dataset. We compare our method to the reference-guided i2i methods Stargan-v2~\cite{starganv} where the target and the source image are from the same domain.}
\label{tab:afhq_metrics}
\vspace{-0.5cm}
\end{table}

\begin{table}[]
\centering
% \resizebox{\linewidth}{!}{%
\begin{tabular}{lcccc}
\toprule
LR target res. & $4\times 4$ & $8\times 8$        & $16\times 16$       & $32\times 32$       \\ 
\midrule
FID$_\downarrow$        & 15.13             & 15.34    & 19.45    & 13.55    \\
LPIPS$_\uparrow$        & 0.30           & 0.34     & 0.14     & 0.08     \\ 
\bottomrule
\end{tabular}%
% }
\caption{Effect of the LR resolution on the FID and LPIPS metrics for the CelebA-HQ dataset.}
\label{tab:resolution}
\vspace{-0.2cm}
\end{table}

%!TEX root = egbib.tex
\section{Discussion}

% Order: +/-/+
% Start by summarizing work, highlighting contributions
% Limitations and future work 

This paper proposes a novel framework for reference-guided image synthesis targeted towards the scenario where the reference (target) has very low resolution. Our method attempts to realistically fuse information from both the high resolution source (such as identity and HR facial features) and the low resolution target (such as overall color distribution and pose). Our experiments show that our method allows for the generation of a wide variety of realistic images given LR targets. We validate our method on two well-known datasets, CelebA-HQ and AFHQ, compare it to the leading i2i methods \cite{starganv, MUNIT, DRIT}, and demonstrate advantages in terms of diversity and visual quality. 
 % and we discussed our limitations in the ablation study on the LR size.   

\paragraph{Limitations and Future Work}
As with recent work on exemplar-based super-resolution~\cite{srntt}, our method works best when the LR and HR images come from the same domain (human faces, for example). In the case where the target LR image comes from a different domain than the source (e.g.\ tiger vs lion), the generated image attempts to match the LR target at the expense of ``forgetting'' more information from the source image. In addition, we also note that results sometimes lack diversity for a given LR target (rows in fig.~\ref{fig:lots}, for example)---this is a consequence of having to perfectly match the LR image.
A potential solution to mitigate both of the above problems would be to soften that constraint, for instance by increasing the distance $\epsilon$ in eq.~\ref{Rx}, or by modifying the discriminator inputs (eq.~\ref{eq:disInput}) to tolerate larger differences. Finally, the framework has so far only been tested on faces (humans and animals). Extending it to handle more generic scenes, where the LR target would capture higher level information such as layout, makes for an exciting future research direction. 

% on closely related images and on very low resolution targets, where it can excel by generating high quality, realistic images. 
% Although, it's possible to apply on domains containing wide varieties, it can hurt the diversity of the model and reduce the similarities between the source image and the generated one, which can be considered as the main limitation of our work. Such issues will be the subject of our future works.   

{\small
\bibliographystyle{ieee_fullname}
\bibliography{egbib}
}

\end{document}